# Adaptive Double-Booking Strategy for Outpatient Scheduling Using Multi-Objective Reinforcement Learning

Ninda Nurseha Amalina, Heungjo An[1]

Department of Industrial Engineering, Kumoh National Institute of Technology, Gumi City, South Korea

**Abstract**

Patient no-shows disrupt outpatient clinic operations, reduce productivity, and may delay necessary care. Clinics often adopt overbooking or double-booking to mitigate these effects. However, poorly calibrated policies can increase congestion and waiting times. Most existing methods rely on fixed heuristics and fail to adapt to real-time scheduling conditions or patient-specific no-show risk. To address these limitations, we propose an adaptive outpatient double-booking framework that integrates individualized no-show prediction with multi-objective reinforcement learning. The scheduling problem is formulated as a Markov decision process, and patient-level no-show probabilities estimated by a Multi-Head Attention Soft Random Forest model are incorporated in the reinforcement learning state. We develop a Multi-Policy Proximal Policy Optimization method equipped with a Multi-Policy Co-Evolution Mechanism. Under this mechanism, we propose a novel $\tau$ rule based on Kullback–Leibler divergence that enables selective knowledge transfer among behaviorally similar policies, improving convergence and expanding the diversity of trade-offs. In addition, SHapley Additive exPlanations is used to interpret both the predicted no-show risk and the agent's scheduling decisions. The proposed framework determines when to single-book, double-book, or reject appointment requests, providing a dynamic and data-driven alternative to conventional outpatient scheduling policies.

**Keywords**

Outpatient Scheduling, Double Booking, No-Show Prediction, Multi Objective Reinforcement Learning, SHAP

## 1. Introduction

A missed scheduled appointment without prior cancellation, commonly referred to as a patient no-show, presents a persistent challenge in outpatient clinic operations. Previous studies have shown that patient no-shows reduce system efficiency, leading to lower clinic productivity, longer waiting times, and decreased patient satisfaction [1], [2], [3]. These operational disruptions may also contribute to adverse clinical outcomes, including delayed diagnoses and interruption in continuity of care. In addition, patients who miss scheduled appointments may subsequently seek urgent or emergency care after losing the opportunity for timely outpatient treatment.

To address the impact of no-shows, healthcare providers frequently implement overbooking strategies by scheduling more patients than the nominal clinic capacity in anticipation that some patients will not attend [4], [5], [6]. Overbooking can reduce idle time and improve slot utilization by ensuring that available appointment slots are not left unused. However, poorly designed overbooking policies may result in overcrowding, extended waiting times, and increased staff workload when actual attendance exceeds expectations. These conditions can negatively affect both patient satisfaction and the quality of care.

---
[1] Correspondence: heungjo.an@kumoh.ac.kr

Within the broader category of overbooking strategies, double-booking, which assigns two patients to a single appointment slot, has gained attention due to its simplicity and widespread use in practice [7]. Although double-booking can mitigate the effects of no-shows, applying this strategy uniformly across all slots without considering patient-specific no-show risk may lead to excessive congestion. Previous research suggests that double-booking is most effective when the probability of a patient no-show is sufficiently high [8]. Therefore, selectively double-booking patients with higher no-show risk is essential to balance operational efficiency and service quality.

Despite the effectiveness of overbooking and double-booking, several gaps remain in the current study. First, most scheduling policies are developed in a static manner, often at the session or daily level, and do not adapt to the evolving sequence of booking requests, cancellations, and capacity changes over a multi-day horizon. Second, although patient no-show prediction models have been incorporated into appointment scheduling frameworks through simulation studies [4], [5], [7], [9], limited attention has been given to real-time adaptive scheduling systems that continuously integrate predictive information into decision-making.

Traditional optimization methods, including stochastic programming and heuristic approaches, have been widely applied to appointment scheduling problems. However, these methods often face limitations in representing the sequential and uncertain nature of outpatient clinic operations, where decisions and outcomes unfold over time. Reinforcement learning (RL) provides an alternative framework that can learn adaptive scheduling policies through repeated interaction with the system environment and feedback from observed outcomes.

While RL has been applied in various healthcare decision-making problems, most existing studies focus on single-objective optimization and do not explicitly address the multiple competing objectives present in outpatient scheduling. Furthermore, to the best of our knowledge, RL has not been applied specifically to overbooking or double-booking decisions in outpatient clinics.

Since overbooking and double-booking are commonly used in practice to manage patient no-shows, integrating these strategies with RL has the potential to improve scheduling decisions by allowing policies to adapt dynamically to patient behavior and system conditions. Such an approach enables scheduling decisions that respond to real-time information rather than relying on fixed rules.

To address this gap, the objective of this study is to develop an adaptive double-booking strategy for outpatient scheduling using a multi-objective RL framework that incorporates individualized no-show predictions into real-time booking decisions. The proposed model decides when to single-book, double-book, or reject appointment requests, aiming to balance three competing objectives by maximizing slot effectiveness, minimizing double-show risk, and aligning expected attendance with one patient per slot. The detailed definitions of three objectives are described in Section 3.1.

This study makes four main contributions. First, we present, to the best of our knowledge, the first multi-objective Markov Decision Process (MDP) formulation of outpatient scheduling problem that explicitly models single-book, double-book, and reject decisions, while restricting each slot to at most two patients to reflect common double-booking practices. Second, we integrate individualized no-show probabilities, predicted using our previously proposed Multi-Head Attention Soft Random Forest (MHASRF) model, directly into the decision-making process. Third, we implement a Multi-Policy PPO (MPPPO) framework with a Multi Policy Co-Evolution Mechanism (MPCEM). In this framework, multiple policies are trained in parallel under different objective weight configurations to approximate a pareto frontier. To improve training stability and trade-off coverage, we evolved the fixed τ approach into a novel KL-based τ mechanism that selectively transfers knowledge among behaviorally similar policies. We also use SHAP (Shapley Additive Explanations) to analyze how the RL agent makes its booking decisions.

Overall, this research presents an adaptive, data-driven scheduling framework designed to enhance the efficiency and reliability of outpatient clinic operations. The proposed framework enables smarter decision-making by determining when to single-book, double-book, or reject appointment requests based on patient-specific predicted no-show probabilities and real-time scheduling information.

The remainder of this paper is organized as follows. Section 2 reviews the related literature. Section 3 describes the problem statement, framework overview, and model components. Section 4 presents the experimental setup, results, and analysis. Finally, Section 5 concludes the study and discusses future work.

The corresponding code implementing the approach is publicly available.[2]

## 2. Literature Review

Patient no-shows, which miss appointments without prior cancellation, result in unused capacity, disrupt the continuity of care, and lead to costly delays as waiting for unpredictable patient attendance. To mitigate these disruptions, clinics often use strategies like overbooking (scheduling more patients than available slots) or double-booking (assigning two patients to the same appointment) [7]. While these approaches can improve resource utilization when some patients fail to attend, they may also lead to overcrowding when all scheduled patients arrive, potentially resulting in longer waiting times and a decrease in service quality [10].

The resulting scheduling challenge is inherently multi-objective, as clinics aim to maximize the use of available appointment slots while minimizing both the likelihood and the negative consequences of overcrowding. Ideally, they strive to match the expected number of patients attending each slot with its intended single-patient capacity. To achieve this, any effective scheduling strategy must address several complexities, including uncertainty in patient attendance, variation in individual patient behavior, and progressive booking decisions over a period of days or even weeks prior to service.

Given these challenges, prior research has explored a range of strategies to manage no-show risk and balance competing objectives in outpatient scheduling. Early studies addressed no-shows using heuristic rules, often validated via discrete-event simulation (DES). Study [11] compared open access and overbooking, demonstrating that modest overbooking can outperform open access by reducing unmet demand and improving slot utilization, while highlighting trade-offs in overtime and patient waiting time. A subsequent DES study [12] examined overbooking schemes under realistic appointment flows and service processes. In addition, study [13] evaluated how time-slot structures — such as fixed-length intervals, dome patterns (where appointment slots are shorter in the middle of the day and longer at the ends), or flexible start times — affect performance under no-show uncertainty. From these studies [11], [12], [13] well-placed overbooking can materially improve throughput, whereas naïve policies may lead to long waits and excessive overtime when attendance exceeds expectations.

Building on these heuristic and simulation-based approaches, subsequent work adopted a more formal mathematical perspective by framing outpatient scheduling as a stochastic optimization problem. Representative models determine the daily number of appointments and their replacement decisions to minimize expected costs associated with patient waiting time, provider idle time, and overtime [10], [14], [15]. Extensions incorporate capacity planning with overbooking using queueing-inspired formulations [16] or booking-limit optimization that accounts for late-cancellations and resource coupling constraints [17]. Study [18] further integrates a stochastic mixed-integer linear program with simulation to explicitly navigate trade-offs between operational efficiency and patient accessibility. These methods provide explicit control of trade-offs between efficiency and accessibility. Overall, these optimization-based approaches offer interpretable solutions and explicit control over efficiency–accessibility trade-offs; however, they often assume well-calibrated distributions. As a result, they may have limited adaptability to sequential appointment arrivals and evolving system state over multi-day scheduling horizons.

---

[2] https://github.com/amalinand/DBRL

In parallel, a growing subset of studies has examined double-booking as a distinct overbooking strategy, often leveraging predictive models and behavioral insights. Within the broader overbooking literature, double-booking has emerged as a focused tactic. Study [7] combined predictive analytics using logistic regression and random forest with hybrid simulation (agent-based and DES) to design and evaluate prediction-based double-booking policies, that demonstrating superior performance compared with random or time-designated rules. Study [19] investigated strategic patient behavior, showing that some customers intentionally double-book across providers to secure faster service and analyzing when to intervene. Collectively, these works demonstrate improvements beyond simple heuristics, but also highlight the need for more adaptive scheduling strategies that determine which patients should be double-booked and when double-booking should be applied within the session.

Classical approaches demonstrate that thoughtfully designed overbooking and double-booking can mitigate no-shows; however, their effectiveness depends on predictive insight and the ability to adapt as bookings accumulate over time. Study [4] used logistic regression and tree-based models to estimate attendance from demographics, appointment context, lead time, and past behavior. They computed overbook-able minutes per day by aggregating confidence-adjusted no-show probabilities across 30-minutes slots. When the estimated overbook-able minutes exceeded the length of one appointment slot, an additional patient was added to the schedule. Study [5] developed a Bayesian nested logit for individualized no-shows and showed that predictive-driven overbooking reduced waiting when no-show risk was low while decreasing idle-time costs when no-show risk was high. Similarly, another study [9] trained prediction models and simulated multiple overbooking levels, finding that prediction-based policies outperformed uniform, percentage-based overbooking rules. These studies indicate that a personalized forecast can substantially improve scheduling decisions relative to population averages or fixed rules.

More recently, study [20] proposed Multi-Head Attention Soft Random Forest (MHASRF), a hybrid model that incorporates attention mechanisms with probabilistic soft splitting to capture complex feature interactions while preserving interpretability. MHASRF achieved well-balanced and strong performance. In this study, MHASRF serves as the predictive layer by providing individualized no-show probabilities that are subsequently used as inputs to the scheduling decision model.

Prediction alone does not prescribe sequential actions. Therefore, clinics must translate attendance forecast into booking decisions while balancing multiple objectives. This gap has motivated optimization or simulation-based decision pipelines. However, in fast moving outpatient settings characterized by stochastic, rolling arrivals, solving a new optimization after every booking can be computationally burdensome, and model misspecification may degrade performance. These limitations motivate the need for learning policies that can adapt online through interaction with the system —an ability naturally supported by RL.

RL is well suited to sequential decision-making problems under uncertainty, particularly when explicit environment models are incomplete, or computationally expensive to solve. In revenue management, airline overbooking was among the earliest applications of RL. Study [21] formulated single-leg seat control with overbooking as a semi-MDP and developed RL algorithms that outperformed classical heuristics. More recently, study [22] employed deep Q-network (DQN) to learn seat inventory and overbooking controls directly from simulation, demonstrating the potential of deep RL in complex stochastic settings.

Inspired by such developments, healthcare has increasingly explored RL for appointment scheduling, including teleconsultation and outpatient settings [23], [24]. In these studies, the booking process is typically modeled as an MDP and value-based methods are used to dynamically allocate appointment slots. However, despite these promising advances, two major limitations remain. First, to our knowledge, RL studies in appointment scheduling do not effectively incorporate individualized no-show probabilities into the state representation. Consequently, learned policies often fail to make fine-grained decisions about which patient to single-book or double-book in which slot as bookings evolve over time. Second, most prior RL-based appointment scheduling approaches optimize a single scalar reward. In contrast, real-world clinics must balance competing objectives, including maximizing slot effectiveness, minimizing double-show risk, and matching expected attendance to slot capacity. To address these limitations, we propose an RL-based double-booking strategy for outpatient scheduling. Specifically, we develop a prediction-aware, multi-objective RL framework that adapts online to evolving conditions and explicitly targets clinically meaningful trade-offs relevant to real-world clinic operations.

## 3. Methodology

### 3.1 Problem Statement: Outpatient Appointment Scheduling

This study addresses the outpatient appointment scheduling problem in which each scheduled appointment is associated with a patient-specific probability of no-show. Missed appointments can lead to underutilized time slots and wasted physician capacity. To mitigate this issue, clinics commonly adopt overbooking strategies by assigning multiple patients to the same appointment slot to reduce idle time. However, overbooking must be applied carefully, as excessive overbooking may cause clinic overcrowding, longer patient waiting times, and increased physician workload.

To balance operational efficiency and service quality, this study focuses on double-booking, which allows at most two patients to be scheduled in the same time slot. Scheduling three or more patients per slot is excluded because it substantially increases the risk of overcrowding and disruption to clinical workflow.

The outpatient scheduling problem is further characterized by inherent uncertainty arising from stochastic appointment requests and individualized no-show probabilities. Patients request appointments with varying lead times, and each patient has a distinct likelihood of attending the scheduled appointment. As a result, scheduling decisions must be adapted dynamically to these uncertainties in order to achieve reliable system performance.

The study considers a set of clinics C. Each clinic $c \in C$ consists of several departments $D_c$, and each department $d \in D_c$ includes multiple physicians $P_{dc}$. Each physician has a predefined number of appointment slots per day, and the scheduling horizon is restricted to a fixed period. With this hierarchical structure across clinics, departments, and physicians, this study focuses on the pre-arrival stage, covering the process from the moment a booking request is made until the scheduled appointment time, when patient attendance (show or no-show) is realized. This stage involves the substantial uncertainty—such as no-shows, variable arrival behavior, and slot-level congestion—which directly affects scheduling performance. Post-arrival processes, including patient waiting times and consultation flow, fall outside of this study and are therefore not modeled in this framework.

When a booking request is received, the scheduler must decide whether to schedule the patient using single-booking or double-booking, or to reject the request if no suitable appointment slot is available. These decisions are made sequentially over time as booking requests arrive, up to the appointment day. The scheduler has three objectives. First, it seeks to maximize effective slot utilization by maintaining high attendance while avoiding wasted capacity caused by no-shows. Second, it endeavors to minimize double-show events—where two patients attend the same time slot—which increases the risk of overcrowding and service disruption. Third, it aims to align the expected attendance per slot with the intended single-patient capacity, thereby ensuring a predictable workload for healthcare providers.

Many prior studies evaluate scheduling performance using slot utilization as the primary metric, implicitly assuming that higher utilization indicates better performance. In outpatient clinics where each appointment slot is designed for a single patient, this assumption may not hold under double booking. A slot can be fully utilized while still being operationally overloaded when both scheduled patients attend, which increases the risk of overcrowding and service disruption. To address this issue, this study adopts effective slot utilization as one of the performance measures.

Figure 1 illustrates the outpatient double-booking scheduling setting, in which patients submit booking requests across different clinics, departments, and appointment times. Each request of appointment is associated with a patient-specific no-show probability, which introduces uncertainty and motivates the need for a dynamically adaptive scheduling approach.

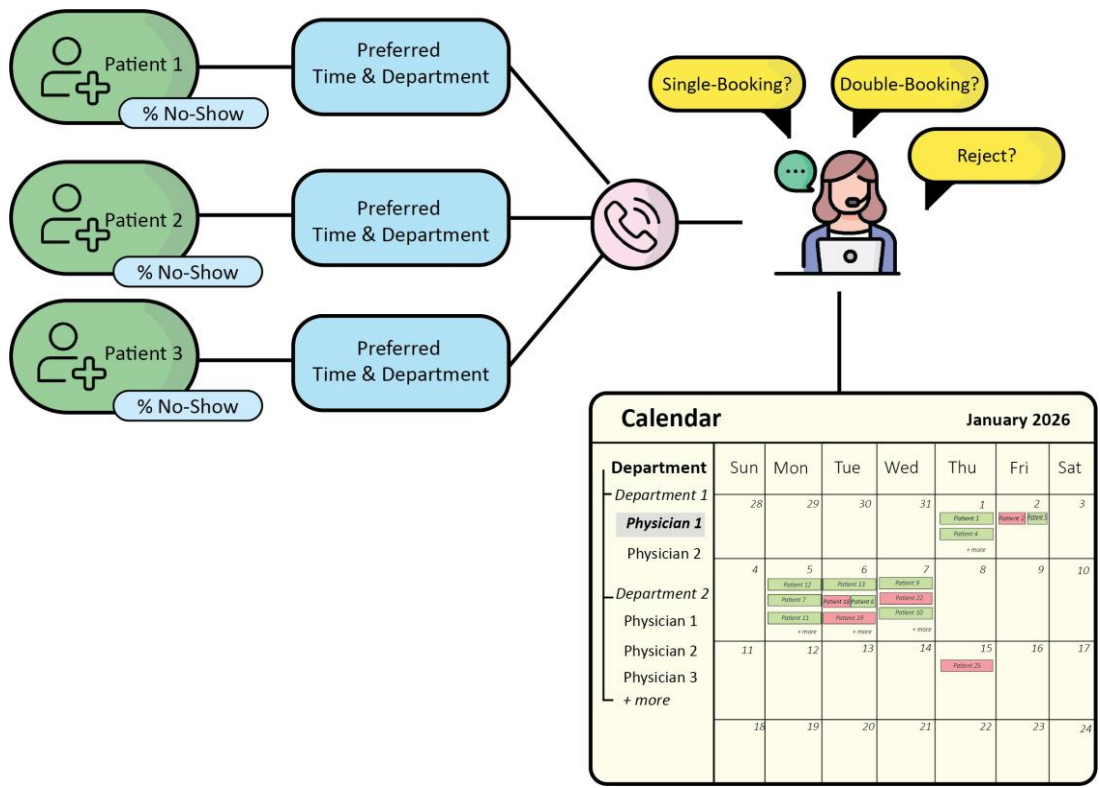

**Figure 1** Outpatient double-booking scheduling setting. Patients submit appointment requests with a preferred department and appointment time, and each request is associated with an individualized no-show probability. Upon receiving each request, the scheduler dynamically decides among single-booking, double-booking, or rejection based on slot feasibility. The calendar illustrates the resulting schedules across multiple departments and physicians over the planning horizon, including both single-booked and double-booked slots. Green blocks represent attending patients (show), while red blocks represent non-attending patients (no-show).

Table 1 presents notations used in the appointment scheduling model, which considers multiple clinics, departments, and physicians.

*Table 1 Description of Notation in Appointment Scheduling*

| Notation | Description |
|---|---|
| **Sets** | |
| $C$ | Set of clinics |
| $D_c$ | Set of departments in clinic c |
| $P_{dc}$ | Set of physicians in department $d$ of clinic c |
| $S_{pdc}$ | Set of appointment slots available for physician $p$ in department $d$ of clinic $c$ |
| $T_n$ | Set of event time steps occurring on day $n$ |
| $T$ | The full set of all event time steps across horizon |
| **Indices** | |
| $c$ | Clinic index, $c \in C$ |
| $d$ | Department index, $d \in D_c$ |
| $p$ | Physician index, $p \in P_{dc}$ |
| $s$ | Slot index, $s \in S_{pdc}$ |
| $n$ | Day index within the scheduling horizon ($n = 1, 2, …, H$) |
| $t$ | Event time steps |
| $i$ | Patient index |
| **Parameters** | |
| $\pi_i$ | Predicted no-show probability for patient $i$ |
| $\lambda$ | Expected number of booking requests per day (Poisson rate) |
| $L_i$ | Lead time of patient $i$ (days between booking and appointment date) |
| $A_i$ | Attendance indicator, $A_i \sim$ Bernoulli $(1-\pi_i)$ |
| $H$ | Length of scheduling horizon (14 days) |

## 3.2 Framework Overview

The proposed framework formulates outpatient appointment scheduling as a sequential decision-making problem, in which each booking decision alters future slot availability and, consequently, affects patient flow, and overall system efficiency. The primary objective is to maximize appointment slot utilization while controlling the overcrowding risk caused by overbooking.

Figure 2 summarizes the overall framework. The framework integrates individualized no-show prediction with adaptive scheduling in a unified workflow. For each incoming booking request, we first estimate the patient-specific no-show probability using the MHASRF model. MHASRF leverages patient-level and appointment-level features to generate a personalized no-show probability, which is subsequently used as an input to the scheduling decision process.

We then model the scheduling environment as an MDP to capture its dynamic and stochastic nature of the scheduling environment. The MDP formulation captures how booking actions influence future states, including remaining slot capacity, double-booking indicators, and other operational conditions.

To derive scheduling policies, we solve the MDP using RL through interactions with a simulated environment. We employ MPPPO combined with a MPCEM [25], [26], as this framework can handle high-dimensional, context-rich state representations while learning a diverse set of policies that specialize in different trade-offs among the three objectives. In addition, MPCEM enables knowledge transfer among policies, improving training stability. A detailed description of the solution algorithms is provided in Section 3.5. At each booking event, the agent selects one of three actions —single-book, double-book, or reject—based on the current system state and the predicted no-show risk.

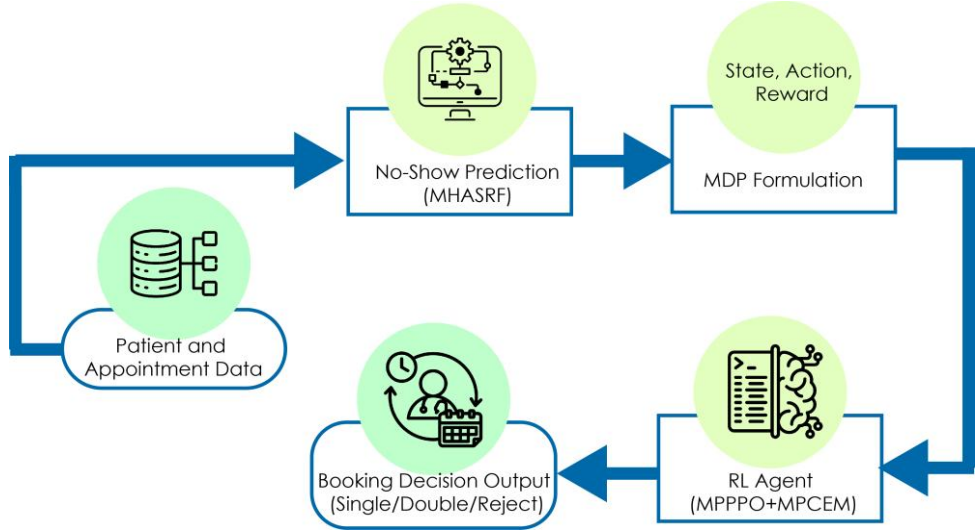

**Figure 2.** Overall framework adaptive double-booking strategy, integrating no-show prediction, MDP formulation, and an RL-based booking policy.

### 3.3 Model Components

#### 3.3.1 No-Show Prediction Model

This study adopts the MHASRF, a no-show prediction model previously proposed and validated by our research group, to estimate patient-specific no-show probabilities. MHASRF combines a soft-splitting random forest with a multi-head attention mechanism that assigns adaptive weights to patient and appointment information, allowing the model to capture heterogeneous patterns in attendance behavior. Unlike conventional tree-based methods that rely on hard decision splits, MHASRF applies probabilistic soft splitting, which provides greater modeling flexibility and supports interpretability. In our prior study, MHASRF achieved consistently strong predictive performance across multiple metrics and outperformed widely used baselines, including decision tree, random forest, logistic regression, and naïve Bayes. For each booking request *i,* MHASRF produces a predicted no-show probability $\pi_i$. This value is incorporated into the scheduling environment state and used by the RL agent to guide booking decisions.

#### 3.3.2 Simulation Environment for Appointment Scheduling

This section focuses on the simulation logic and event-driven dynamics of the environment, while the formal MDP formulation is provided in the next section. The RL environment is implemented as a DES with two event types: booking events and appointment arrival events. Each episode covers a 14-day planning horizon, during which all events are processed in chronological order.

During booking event, patient requests arrive according to a Poisson process with rate $\lambda$. For each request, patient attributes are generated and the individualized no-show probability $\pi_i$ is predicted using MHASRF. A physician is selected using a load-balancing assignment rule. The lead time $L_i$ is then sampled from a Gamma distribution to determine the patient's appointment day, calculated as $n_i = n_t + L_i$, where $n_t$ is the current booking day. The resulting appointment day is clipped to remain within the valid scheduling horizon of $H$ days. The RL agent observes the current system state $s_t$ and selects an action $a_t$ according to the policy being learned. After the action is selected, a shaped reward is computed, and the booking request is recorded. The system state is then updated based on the chosen action.

For each incoming request, the environment identifies feasible appointment slots using hierarchical search procedure to determine feasible scheduling options before the agent selects an action. The search first attempts to schedule the patient in the requested slot. If that slot is unavailable, the environment searches alternative days with the horizon. If no suitable slot exists for the initially assigned physician, the search is expanded to other physicians within the same department. Once a feasible slot is identified, the agent chooses an action $a_t \in \{0, 1, 2\}$, where $a_t = 0$ indicates single-book the patient, $a_t = 1$ indicates double-book the patient, and $a_t = 2$ indicates rejecting the request when no feasible slots remain.

When the appointment arrival event occurs, meaning the appointment day is reached, the patient's attendance is simulated as a Bernoulli random variable $A_i \sim \text{Bernoulli}(1 - \pi_i)$. The environment updates the slot status (show or no-show) and applies the realized rewards based on actual attendance. A simulation flow is illustrated in Figure 3.

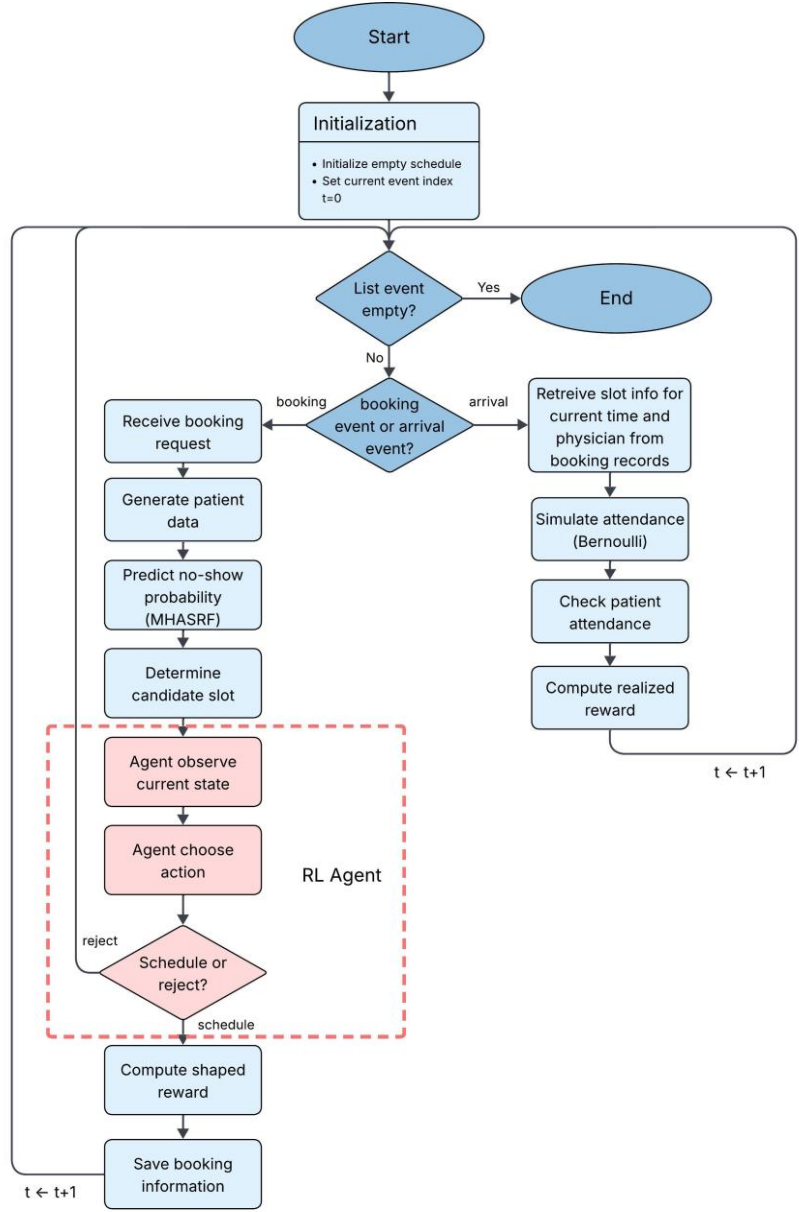

**Figure 3** summarizes the workflow of the simulation environment. At each time step, the environment processes the next event in chronological order, which is either a booking event or an appointment arrival event. During a booking event, the environment generates patient attributes, predicts the no-show probability using MHASRF, identifies feasible appointment slots, and provides the current state to the RL agent. The agent then selects one of three actions, single-book, double-book, or reject, after which the environment records the decision, updates the schedule, and computes the shaped reward. During an arrival event, the environment simulates patient attendance using a Bernoulli trial and computes the realized reward based on whether the patients shows or no-shows.

To maintain a tractable scheduling environment while preserving essential operational features, the model adopts the following assumptions:

a. Each physician provides service for a fixed number of time slots per day (from 8:00 AM to 4:00 PM at 30-minute intervals).
b. Patient arrivals for booking requests follow a Poisson process with rate λ.
c. Patient attendance on the appointment day follows a Bernoulli distribution based on the predicted no-show probability.
d. Patient lead time on the appointment request to appointment day follows a Gamma distribution.

These assumptions are commonly adopted in previous studies on overbooking and appointment scheduling models [10].

### 3.4 Markov Decision Process (MDP) Formulation

We formulate the outpatient appointment scheduling problem as an MDP defined by a state space, action space, transition dynamics, and reward function.

- **State**

The state represents the scheduling environment at a booking decision epoch and summarizes relevant operational information. The state is defined as:

$$s_t = \{c, d, p, n, s, \omega, \pi_i, \zeta, N_p, A_p\} \quad (2)$$

where $c$, $d$, and $p$ denote the clinic, department, and physician associated with the candidate slot for the current request; $n$ and $s$ denote the appointment date and time slot; $\omega$ denotes the slot status, which can be available, single-booked, or double-booked; $\pi_i$ is the predicted no-show probability for the current patient $i$; $\zeta$ indicates whether the slot remains eligible for double-booking; $N_p$ is the number of patients currently scheduled with the physician; and $A_p$ denotes the remaining available slots of physician $p$. The state representation is designed to satisfy the Markov property by fully capturing all information required to determine future slot availability, physician workload, and expected attendance outcomes.

- **Action**

At each booking event $t$, the agent selects an action:

$$a_t \in \{0, 1, 2\} \quad (3)$$

Where $a_t = 0$ corresponds to single-booking, $a_t = 1$ corresponds to double-booking, and $a_t = 2$ corresponds to rejecting the booking request when no feasible slot is available. This action space allows the agent to balance slot utilization and congestion risk.

- **Transition dynamics**

State transitions occur through two types of events: booking events and appointment arrival events. The agent takes actions only at booking events, while appointment arrival events serve as exogenous transitions that reveal stochastic outcomes without agent intervention. During booking events, the selected action updates the scheduling configuration by modifying slot occupancy, remaining capacity, and physician workload. The subsequent state reflects the updated system and the next incoming booking request. The transition does not correspond to the progression of a single patient but changes in the overall scheduling environment.

During appointment arrival events, no action is taken. Instead, patient attendance is realized stochastically based on the predicted no-show probability $\pi_i$. Arrival events reveal the outcomes of earlier booking decisions and update slot outcomes accordingly. Although patients may arrive early or late, the current RL model abstracts away such arrival-time variations and evaluates outcomes solely based on show versus no-show events.

- **Reward**

The reward function is multi-objective and captures the key trade-offs in outpatient scheduling. At each decision epoch, the immediate reward is defined as:

$$R(s_t, a_t) = \alpha U_t + \beta D_t + \gamma B_t \tag{4}$$

Where $U_t$ represents effective slot utilization, $D_t$ represents double-show avoidance, and $B_t$ represents attendance balance. The weight α, β, and γ reflect the relative importance of these objectives. However, the appointment outcomes are observed only on the appointment day which makes rewards inherently delayed. To improve learning efficiency, we incorporate a shaped reward based on predicted no-show probabilities [27], which provides an expectation-based estimate of future performance prior to observing realized attendance.

- **Realized reward**

After appointment outcomes are observed, realized reward are computed as:

$$U_t = 1 - |S_t - 1| \tag{5}$$

$$D_t = \begin{cases} 1, & \text{if only one patients in a double-booked slot showed up} \\ 0, & \text{if both patients in a double-booked slot showed up} \end{cases} \tag{6}$$

$$B_t = \begin{cases} \max(0, 1 - |E_t - 1|), & \text{if at least one patient scheduled} \\ 0, & \text{if no patient scheduled in the slot} \end{cases} \tag{7}$$

Where $S_t$ is the number of attending patients in slot t, and Et is the expected attendance based on predicted no-show probabilities $\pi_i$. Expected attendance is computed as:

$$E_{t,SB} = 1 - \pi_i \tag{8}$$

$$E_{t,DB} = (1 - \pi_{i_1}) + (1 - \pi_{i_2}) \tag{9}$$

- **Shaped reward**

The shaped reward provides early feedback based on expected attendance:

$$U_{t(shaped)} = \begin{cases} 1 - \pi_i, & \text{if single-booked} \\ (1 - \pi_{i_1})\pi_{i_2} + \pi_{i_1}(1 - \pi_{i_2}), & \text{if double-booked} \end{cases} \tag{10}$$

$$D_{t(shaped)} = \begin{cases} 1, & \text{if single-booked} \\ 1 - (1 - \pi_{i_1})(1 - \pi_{i_2}), & \text{if double-booked} \end{cases} \tag{11}$$

$B_{t(shaped)}$ is equivalent to the realized attendance balance component in Eq 7, since it already incorporates expected attendance from the no-show predictions. By providing expectation-based feedback, the shaped reward encourages actions with higher expected utilization and lower congestion risk before outcomes are realized. The shaped reward is not intended as a potential-based transformation but as an auxiliary learning signal to accelerate policy convergence under delayed feedback. Policy performance is ultimately evaluated using realized rewards based on observed attendance outcomes, while the shaped reward is used only during training to mitigate delayed feedback arising from appointment outcomes that are observed only on the appointment day.

### 3.5 Reinforcement Learning Algorithm

Outpatient appointment scheduling is a sequential decision problem under uncertainty, where booking decisions influence future slot availability, congestion risk, and cumulative clinic performance. The system is stochastic due to random booking arrivals, heterogeneous lead times, and patient-specific no-show behavior. In the setting, transition dynamics and realized rewards cannot be specified explicitly in advance. We therefore adopt a model-free RL approach that learns scheduling policies directly from interaction with the environment.

Among model-free methods, policy-based approaches are well suited to this problem due to the high-dimensional and context-rich state representation, which includes clinic and provider attributes, slot occupancy, predicted no-show probabilities, and remaining capacity. Value-based methods such as Q-learning or Deep Q-Networks tend to struggle in such settings because action-value approximate becomes unstable as the state space grows. For these reasons, we adopt Proximal Policy Optimization (PPO) as the foundation of our scheduling framework. PPO improves training stability by updating the policy using a clipped surrogate objective that limits large policy changes. Specifically, let

$$r_t(\theta) = \frac{\pi_\theta(a_t|s_t)}{\pi_{\theta_{old}}(a_t|s_t)}$$

denote the probability ratio, and let $\hat{A}_t$ denote the advantage estimate. PPO updates $\theta$ by maximizing:

$$L^{PPO}(\theta) = \mathbb{E}_t\left[\min\left(r_t(\theta)\hat{A}_t, clip(r_t(\theta), 1-\epsilon, 1+\epsilon)\hat{A}_t\right)\right] \tag{32}$$

where $\epsilon$ is a clipping parameter. PPO is implemented using an actor-critic architecture, where the actor network outputs a probability distribution over the three actions (single-book, double-book, reject) and the critic network estimates state values used for advantage computation.

Standard PPO optimizes a single policy under fixed reward weights, which is limiting in appointment scheduling where multiple competing objectives must be balanced. These objectives include effective slot utilization, avoidance of double shows, and attendance balance. Their relative importance may vary depending on operational priorities. To address this limitation, we implement MPPPO [26], which trains a set of policies in parallel:

$$\Pi = \left\{\pi_{\theta_1}, \pi_{\theta_2}, \ldots, \pi_{\theta_p}\right\} \tag{12}$$

Each policy $\pi_{\theta_p}$ is optimized with a distinct weight vector $w_p = [\alpha_p, \beta_p, \gamma_p]$ and objective:

$$J(\theta_p) = \mathbb{E}_{\pi_{\theta_p}}\left[\sum_{t=0}^{T}\gamma^t\left(\alpha_p U_t + \beta_p D_t + \gamma_p B_t\right)\right] \tag{13}$$

MPPPO extends PPO by training multiple actor-critic pairs in parallel under different objective-weight configurations, allowing each policy to specialize in a different trade-off among the objectives. The resulting policy set provides an approximation of the Pareto frontier, enabling decision-makers to select a scheduling policy aligned with current operational goals.

Training multiple policies independently may lead to suboptimal convergence, as each policy can become trapped in local optimum. To improve learning efficiency and solution diversity, we incorporate a MPCEM presented in [26].

MPCEM enables periodic knowledge transfer among neighboring policies in weight space. Importantly, MPCEM does not replace PPO updates but serves as an auxiliary mechanism applied periodically to encourage cooperative learning. Let $B(w_p)$ denote the set of neighboring weight vectors closest to $w_p$ and $B(\pi_{\theta_p})$ the corresponding policies. The co-evolution update is defined as:

$$\theta_p \leftarrow \tau \theta_{p_i} + (1-\tau)\theta_p \tag{14}$$

where $\tau \in [0,1]$ controls the strength of parameter sharing.

Using a fixed $\tau$ assumes that all neighboring policies provide equally informative guidance, which may not hold in practice. Policies with similar objective weights can exhibit different behaviors due to exploration history or local optima. To overcome this limitation, we propose a novel adaptive $\tau$ mechanism in this study, in which the degree of parameter sharing is explicitly modulated by behavioral similarity between policies. Behavioral similarity is measured using the Kullback-Leibler divergence between policy action distributions:

$$KL\left(\pi_{\theta_p} \parallel \pi_{\theta_q}\right) = \frac{1}{N}\sum_{n=1}^{N}\sum_{a \in A} \pi_{\theta_p}(a|s_n) \log \frac{\pi_{\theta_p}(a|s_n)}{\pi_{\theta_q}(a|s_n)} \tag{15}$$

The adaptive update coefficient is defined as:

$$\tau_{pq} = \tau_{max} \cdot e^{-\varphi KL\left(\pi_{\theta_p} \parallel \pi_{\theta_q}\right)} \tag{16}$$

where $\tau_{max}$ is the maximum transfer rate and $\varphi$ controls sensitivity to behavioral dissimilarity. The update rule becomes:

$$\theta_p \leftarrow \tau_{pq}\theta_q + \left(1-\tau_{pq}\right)\theta_p \tag{17}$$

This mechanism allows stronger knowledge transfer between behaviorally similar policies while preserving diversity among dissimilar ones. Figure 4 illustrates the full MPPPO framework with adaptive $\tau$ MPCEM. In the MPPPO component (right side), multiple policy and value networks are trained in parallel under different objective weight vectors to learn a diverse set of trade-offs in scheduling decisions. Each policy interacts with the environment, collects trajectories, computes advantage estimates, discounted returns, and updates its parameter using PPO. Since independently trained policies may converge to local optima, we incorporate the adaptive $\tau$ MPCEM module (left side) to enable cooperative learning through periodic knowledge transfer.

The MPCEM procedure is executed every $C$ training step, after several MPPPO updates, to allow information exchange among policies. At each MPCEM phase, the objective values of all policies are evaluated and neighborhood relations are determined in weight space. For each policy, candidate neighbors in the set $B$ are compared and behavioral similarity is measured using the KL divergence between their action distributions. The transfer coefficient $\tau$ is the then adapted based on this similarity, and policy parameters are updated accordingly.

This design encourages stronger transfer among behaviorally similar policies while maintaining diversity among dissimilar policies, improving both convergence and coverage of the trade-off surface.

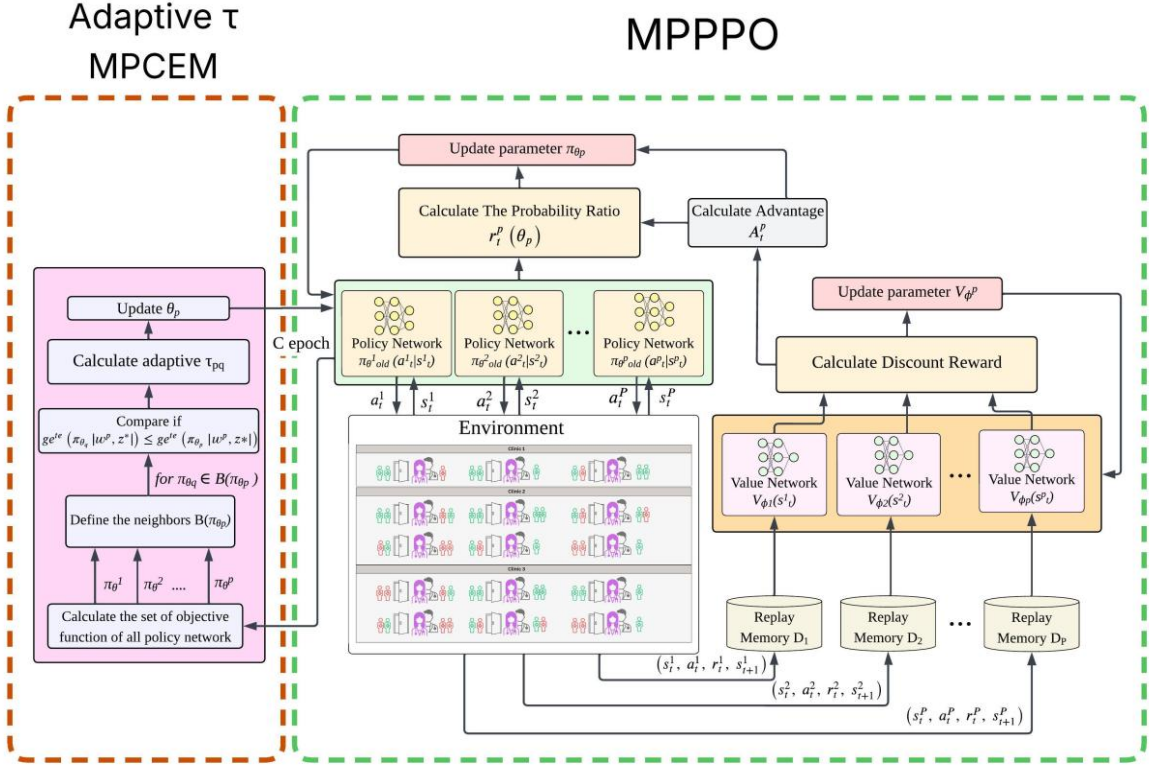

**Figure 4. Overview of MPPPO with adaptive τ MPCEM**. Multiple actor-critic policies are trained in parallel using PPO under different objective-weight configurations (right), while the co-evolution module periodically performs similarity-based parameter sharing among neighboring policies using adaptive τ (left).

### 3.6 Evaluation Metrics

The proposed method is evaluated using four performance metrics:

- Effective slot utilization $\bar{U}$ (%) measures the frequency with which a time slot achieves its intended occupancy of exactly one attending patient. This metric is maximized when a single patient attends the slot, reflecting neither underutilization (zero shows) nor overload (multiple shows). We compute $\bar{U}$ as the average of the utilization component in Eq. 5 with higher values indicating more effective use of slot capacity.

- Double-show avoidance $\bar{D}$ (%) assesses congestion risk under double-booking by measuring how often a double-booked slot avoids a double show. It is computed as the average of the double-show avoidance component in Eq. 6 over double-booked slots only, since a double show cannot occur in single-booked slots. Higher values indicate fewer instances in which both scheduled patients attend simultaneously.

- Attendance balance $\bar{B}$ (%) evaluates how closely expected attendance aligns with the target capacity of one patient per slot, based on predicted no-show probabilities. We compute $\bar{B}$ as the average of the attendance balance component in Eq. 7. Higher values indicate that the policy allocates bookings such that expected attendance per slot remains close to one, thereby balancing utilization and congestion risk.

- Average weighted reward $\bar{R}_p$ is used to assess overall performance within a given objective weight vector. Higher $\bar{R}$ indicates better overall performance under the corresponding objective weight vector.

$$\bar{R}_p = \frac{1}{|T|} \sum_{t \in T} (\alpha_p U_t + \beta_p D_t + \gamma_p B_t) \qquad (18)$$

where $T$ is a set of all appointment slots in the evaluation horizon.

## 4. Experiments and Results
### 4.1 Data and Environment Setup

The experiments use historical outpatient appointment records from a large healthcare provider in the Middle East. For confidentiality reasons, detailed information about the organization is disclosed. The dataset covers January 2018 through December 2018 and contains 157,494 appointment records. Each appointment record contains detailed information used for both prediction and simulation. As summarized in Table 2, the dataset includes patient characteristics, appointment characteristics, clinic and provider information, and external factors. The appointment outcome is recorded as a binary indicator (show or no-show) and is used to model attendance behavior in the simulation environment. The dataset contains both raw features recorded and derived features computed from historical attendance patterns. After data cleaning and preprocessing, 101,532 appointment records were retained for analysis and simulation.

*Table 2 Detailed Description of Variables in the Outpatient Appointment Dataset*

| Features | Description | Type | Source |
|---|---|---|---|
| **Patient Characteristic** | | | |
| Age | Patient age at the scheduled appointment | Continuous | Raw |
| Language | Patient preferred language | Categorical | Raw |
| Gender | Patient gender | Categorical | Raw |
| Visit Reason | The reason for the patient's visit | Categorical | Raw |
| **Appointment Characteristic** | | | |
| Visit Type | The type of visit (e.g., procedure visit, consult visit, new visit) | Categorical | Raw |
| Appointment Status | The current status of the appointment (e.g., no-show, show) | Categorical | Raw |
| Time Appointment by Time | The exact time of the appointment | Continues | Raw |
| Time Appointment by Day | The specific day of the appointment is scheduled | Categorical | Raw |
| Time Appointment by Month | The month in which the appointment is scheduled | Categorical | Raw |
| Week of The Month | The week number within the month of the appointment | Categorical | Derived |
| Season | The season during which the appointment takes place (e.g., summer, winter) | Categorical | Derived |
| Number of Visits | Total number of prior visits before the appointment date/time | Continues | Derived |
| %No-Show | Proportion of previous appointments marked as no-shows | Continues | Derived |
| Number of Appointments on The Same Day | Total same-day appointments for the same patient | Continues | Derived |
| **Clinic & Provider** | | | |
| Institute | The healthcare institute where the appointment is scheduled | Categorical | Raw |
| Center Name | The specific center within the institute where the appointment is scheduled | Categorical | Raw |
| Department Name | The department where the patient is visiting (e.g., dentistry, gynecology, urology) | Categorical | Raw |
| Provider Name | The name of the healthcare provider (physician) assigned to the patient | Categorical | Raw |
| **External Factors** | | | |
| Temperature | The temperature at the time of the appointment | Continues | Derived |
| Dew | The dew point (moisture level in the air) during the time of the appointment | Continues | Derived |
| Humidity | The humidity level at the time of the appointment | Continues | Derived |
| **Windspeed** | The wind speed at the time of the appointment | Continues | Derived |
| **Visibility** | The visibility level during the time of the appointment | Continues | Derived |
| **Weather Conditions** | General description of weather (e.g., raining, cloudy, clear) | Categorical | Derived |
| **Air Quality** | The description of the air quality index (e.g., good, moderate, unhealthy, hazardous) | Categorical | Derived |

The dataset represents a large-scale outpatient system involving multiple clinics, departments, and physicians. To construct a realistic simulation environment, the number of physicians per department is approximated using the average values observed in the data. This approach ensures that simulated capacity and patient allocation patterns are consistent with empirical operating conditions.

Individualized patient no-show probabilities $\pi_i$ are generated using MHASRF. Given its strong predictive performance, MHASRF provides reliable estimates of patient attendance behavior. These predicted probabilities are incorporated into the simulation environment to model heterogeneous patient behavior during booking and arrival events.

The appointment scheduling problem is implemented as a custom OpenAI Gym environment that simulates sequential booking and appointment arrival events. Each simulation episode spans a fixed 14-day planning horizon. This horizon design reflects the practical setting in which outpatient appointments are commonly scheduled about one week in advance [4], while ensuring that the outcomes of booking decisions can be fully observed within the same episode. Specifically, patients may request appointments at any time during episode and are scheduled into future slots based on their lead time. If the horizon were limited to only seven days, many appointments scheduled near the end of the episode would fall outside the simulation window. By extending the horizon to 14 days, the environment captures both the booking process and the subsequent realization of scheduled appointments, while maintaining computational tractability.

### 4.2 Experimental Design

All numerical experiments are implemented in Python 3.12 and conducted on a Windows 11 system equipped with a 12[th] Gen Intel(R) Core (TM) i7-12700 processor. Training is performed for 250 epochs, with five episodes per epoch. Within each episode, booking requests arrive according to a Poisson process with an average rate of 100 requests per day. The main training hyperparameters are summarized in Table 3, and both the actor and critic are implemented using multilayer perceptrons (MLPs).

*Table 3 Training Algorithm Parameter Settings*

| Parameters | Value | Description |
|---|---|---|
| Number of training epochs | 250 | Total number of policy update iterations |
| Number of Policy Networks | 10 | Number of policies trained simultaneously |
| Control parameter of MPCEM | 10 | Controls frequency of MPCEM update |
| Soft update coefficient (max) | 0.5 | Maximum parameter blending factor during MPCEM |
| KL-divergence decay coefficient | 0.5 | Controls sensitivity of adaptive tau |
| Neighbors set size | 2 | Number of nearest policies considered for co-evolution |
| Actor learning rate | 0.0003 | Learning rate for actor network |
| Critic learning rate | 0.0003 | Learning rate for critic network |
| Discount factor | 0.99 | Reward discount factor |
| # hidden layers in MLP | 2 | Hidden layer for each actor and critic network with 128 nodes |

To capture heterogeneous operational priorities, ten policies are trained simultaneously under the MPPPO scheme using different combinations of objective weights ($\alpha$, $\beta$, $\gamma$), which respectively weight effective slot utilization ($U$), double-show avoidance ($D$), and attendance balance (B). These weight vectors are selected to span a range of trade-offs. For example, MPPPO 1 assigns full weight to effective slot utilization, whereas MPPPO 9 emphasizes double-show avoidance with reduced emphasis on the other objectives. In contrast, MPPPO 7 distributes the weights approximately evenly across all three objectives. The complete set of weight combinations is reported in Table 4.

*Table 4 Weight Combinations for Different Policies*

| Policy | $\alpha$ | $\beta$ | $\gamma$ |
|---|---|---|---|
| MPPPO 1 | 1 | 0 | 0 |
| MPPPO 2 | 0 | 1 | 0 |
| MPPPO 3 | 0 | 0 | 1 |
| MPPPO 4 | 0.5 | 0.25 | 0.25 |
| MPPPO 5 | 0.25 | 0.5 | 0.25 |
| MPPPO 6 | 0.25 | 0.25 | 0.5 |
| MPPPO 7 | 0.33 | 0.33 | 0.34 |
| MPPPO 8 | 0.7 | 0.2 | 0.1 |
| MPPPO 9 | 0.2 | 0.7 | 0.1 |
| MPPPO 10 | 0.2 | 0.1 | 0.7 |

### 4.3 Result and Analysis
#### 4.3.1 Training Performance

Training performance is evaluated by examining learning trajectories of the total reward. Figure 5 presents the total reward trajectories for all ten MPPPO policies during training. Across policies, total reward increases rapidly during the early training phase, rising from approximately 2.0 to above 2.3 within the first 30–50 epochs. After this initial phase, most policies converge to stable reward levels, indicating consistent learning behavior and convergence across different objective weight configurations.

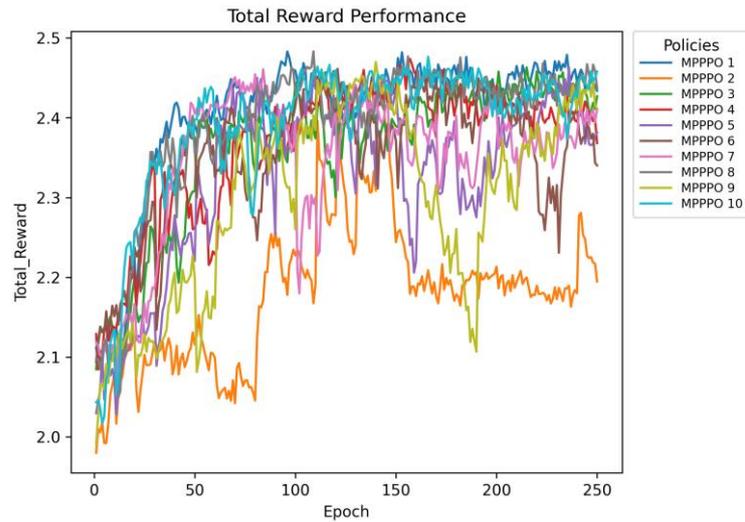

**Figure 5**. **Total Reward Training Performance.** Total reward trajectories of ten MPPPO policies across training epochs, showing a rapid early increase followed by stable convergence.

### 4.3.2 Interpretability Using SHAP

To interpret the learned booking decisions, SHAP are applied to the RL policies. SHAP quantifies the contribution of each state variable to the selection of a given action, thereby revealing which factors most strongly influence the agent's preference for single-booking (action 0) or double-booking (action 1). Analysis focuses on these two actions, as rejection primarily occurs when no slots are available and therefore provides limited behavioral insight. Figure 6 presents SHAP summary plots for both actions.

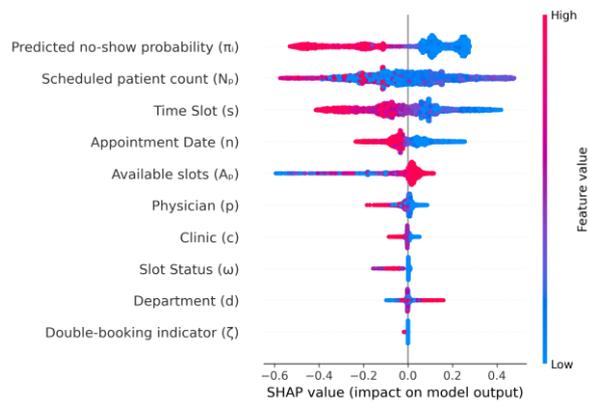

**(a)** SHAP Plot for Single-Booking

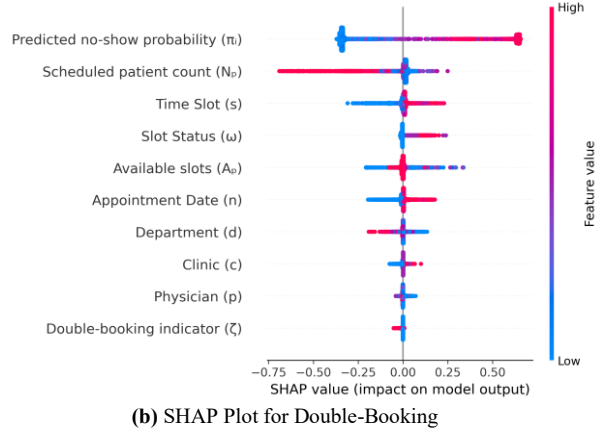

**(b)** SHAP Plot for Double-Booking

**Figure 6. SHAP Plot Summary**. SHAP values indicate the contribution of each state variable to the RL policy's preference for (a) SHAP Plot for Single-Booking Decision, and (b) SHAP Plot for Double-Booking Decision.

For the single-booking decision (Figure 6a), lower predicted no-show probability and lower scheduled patient count strongly increase the likelihood of selecting single-booking. Earlier appointment slots and shorter lead times (sooner appointment dates) also contribute positively, indicating a conservative strategy when appointments are approaching. Overall, patterns suggest that the agent prioritizes operational safety and workload stability, selecting single-booking more often when risk and congestion are low.

For the double-booking decisions (Figure 6b), higher predicted no-show probability is the most influential factor, strongly increase the likelihood of double-booking. In contrast, higher scheduled patient counts reduce the tendency to double-book, suggesting the agent avoids adding risk when workload is already high. Later appointment date slots and greater slot availability also contribute positively, indicating that double-booking is favored when there is flexibility in the schedule. Slot status provides additional context on the current occupancy of the target slot, and the model tends to favor double-booking particularly in partially occupied slot, consistent with adding a second patient when no-show risk is elevated. Overall, these results align with the intended role of double-booking, which is applied selectively under higher uncertainty and tighter capacity constraints.

### 4.3.3 Performance Comparison with Baseline Policies

The trained RL agents under different policies are further evaluated over five test episodes, each simulating a 14-day horizon with booking requests arriving according to a Poisson process at a rate of 100 per day. Their performance is compared against heuristic baseline policies, including a single-booking policy and fixed double-booking policies that apply double-booking when the predicted no-show probability exceeds thresholds of 0.5, 0.6, 0.7, 0.8, and 0.9. These baselines are included because they represent commonly used decision rules in practice and in prior prediction-based scheduling studies, providing an interpretable benchmark to assess whether the proposed RL approach yields improvements beyond fixed heuristics. The threshold values are selected to span conservative to aggressive double-booking behaviors, enabling a fair comparison across different risk preferences.

Results are summarized in Table 5. The MPPPO policies consistently outperform heuristic baselines across most performance metrics. Effective slot utilization for MPPPO ranges from 0.762 to 0.793, compared with 0.678 to 0.706 for fixed double-booking (DB) policies and 0.642 for single-booking (SB) policy. The highest utilization is achieved by MPPPO 10.

Fixed double-booking (DB) policies achieve slightly higher double-show avoidance than MPPPO due to conservative thresholding. However, this improvement comes at the cost of lower utilization and attendance balance. Single-booking (SB) policy achieves the lowest attendance balance, primarily due to frequent underutilization.

MPPPO policies achieve higher attendance balance by dynamically pairing patients to better align expected attendance with slot capacity. This adaptive behavior leads to superior average weighted-reward performance. Reward performance for MPPPO range from 7,119.2 to 8,992.2, compared with 7,006.8 for single-booking and 7,657.6 to 8,058.1 for fixed double-booking. MPPPO 3 achieves the highest reward, followed closely by MPPPO 10 and MPPPO 8. Figure 7 provides a visual summary of the average weighted reward results reported in Table 5, highlighting the relative performance differences across policies.

*Table 5* Average and Standard Deviation of Performance Baseline Comparison

| Policy | Booking Request | Patient Scheduled | Patient Show | Patient No-Show | Effective Slot Utilization ($\bar{U}$) | Double Show Avoidance ($\bar{D}$) | Attendance Balance ($\bar{B}$) | Average Weighted Reward ($\bar{R}_p$) |
|---|---|---|---|---|---|---|---|---|
| MPPPO 1 | 1378.2±19.4 | 1378.2±19.4 | 827.2±22.4 | 551.0±19.6 | 0.783±0.013 | 0.928±0.007 | 0.750±0.009 | 8825.1±120.6 |
| MPPPO 2 | 1378.2±19.4 | 1378.2±19.4 | 818.4±22.8 | 559.8±8.0 | 0.635±0.017 | 0.938±0.003 | 0.615±0.019 | 7119.2±154.9 |
| MPPPO 3 | 1378.2±19.4 | 1378.2±19.4 | 830.4±18.2 | 547.8±19.1 | 0.789±0.016 | 0.937±0.009 | 0.772±0.012 | 8922.2±48.4 |
| MPPPO 4 | 1378.2±19.4 | 1378.2±19.4 | 853.4±14.6 | 524.8±15.7 | 0.791±0.011 | 0.920±0.008 | 0.751±0.009 | 8861.9±81.3 |
| MPPPO 5 | 1378.2±19.4 | 1378.2±19.4 | 838.2±9.4 | 540.0±19.3 | 0.769±0.013 | 0.909±0.011 | 0.717±0.016 | 8759.9±124.3 |
| MPPPO 6 | 1378.2±19.4 | 1378.2±19.4 | 835.6±12.6 | 542.6±23.3 | 0.762±0.008 | 0.909±0.006 | 0.720±0.014 | 8523.5±68.1 |
| MPPPO 7 | 1378.2±19.4 | 1378.2±19.4 | 829.2±16.5 | 549.0±27.5 | 0.773±0.022 | 0.917±0.007 | 0.736±0.016 | 8781.2±77.8 |
| MPPPO 8 | 1378.2±19.4 | 1378.2±19.4 | 827.0±12.5 | 551.2±14.3 | 0.787±0.009 | 0.929±0.005 | 0.751±0.017 | 8874.6±71.5 |
| MPPPO 9 | 1378.2±19.4 | 1378.2±19.4 | 824.6±17.0 | 553.6±5.6 | 0.768±0.009 | 0.922±0.012 | 0.737±0.018 | 8687.3±78.0 |
| MPPPO 10 | 1378.2±19.4 | 1378.2±19.4 | 824.2±14.9 | 554.0±21.9 | 0.793±0.011 | 0.942±0.005 | 0.767±0.013 | 8916.8±86.7 |
| SB | 1378.2±19.4 | 1378.2±19.4 | 849.4±18.2 | 528.8±19.1 | 0.642±0.011 | - | 0.616±0.008 | 7006.8±50.6 |
| DB ≥ 0.5 | 1378.2±19.4 | 1378.2±19.4 | 820.8±15.6 | 557.4 ± 6.1 | 0.692±0.007 | 0.938±0.005 | 0.685±0.009 | 8058.1±113.0 |
| DB ≥ 0.6 | 1378.2±19.4 | 1378.2±19.4 | 826.2±16.5 | 552.0±20.8 | 0.696±0.015 | 0.936±0.008 | 0.679±0.016 | 8005.2 ±50.8 |
| DB ≥ 0.7 | 1378.2±19.4 | 1378.2±19.4 | 818.8±27.6 | 559.4±13.7 | 0.689±0.015 | 0.937±0.009 | 0.680±0.023 | 8022.8±140.1 |
| DB ≥ 0.8 | 1378.2±19.4 | 1378.2±19.4 | 845.0±14.6 | 533.2±9.2 | 0.706±0.010 | 0.936±0.006 | 0.687±0.011 | 8037.4±82.9 |
| DB ≥ 0.9 | 1378.2±19.4 | 1378.2±19.4 | 830.6±11.9 | 547.6±25.0 | 0.678±0.011 | 0.944±0.003 | 0.657±0.015 | 7657.6±57.3 |

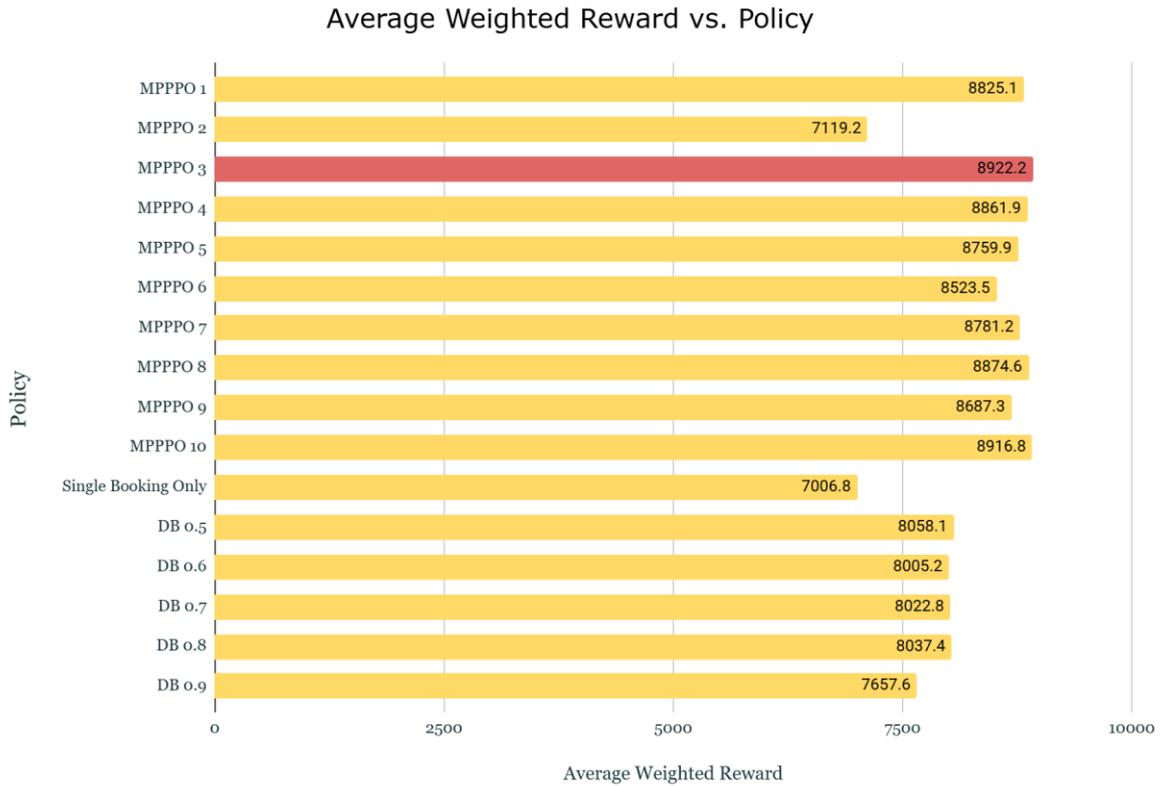

**Figure 7. Average Reward Comparison Across Policies and Baselines.** Average weighted rewards for MPPPO policies and baseline single-booking and fixed double-booking strategies.

### 4.3.4 Trade-Offs Analysis

Figure 8 illustrates trade-offs among effective slot utilization, double-show avoidance, and attendance balance across the ten trained policies. Two policies, MPPPO 3 and MPPPO 10, form an approximate Pareto front, as they are not dominated by any other solutions across all objectives. Both policies assign relatively higher weight to the attendance balance objective.

Across policies, higher utilization or attendance balance is typically associated with lower double-show avoidance, reflecting the inherent trade-off between efficiency and congestion control. In contrast, utilization and attendance balance tend to improve together, indicating that stabilizing expected attendance around one patient per slot also supports effective slot utilization. This pattern suggests that attendance balance provides a strong guiding objective for learning scheduling decision that jointly improve utilization while controlling overbooking risk.

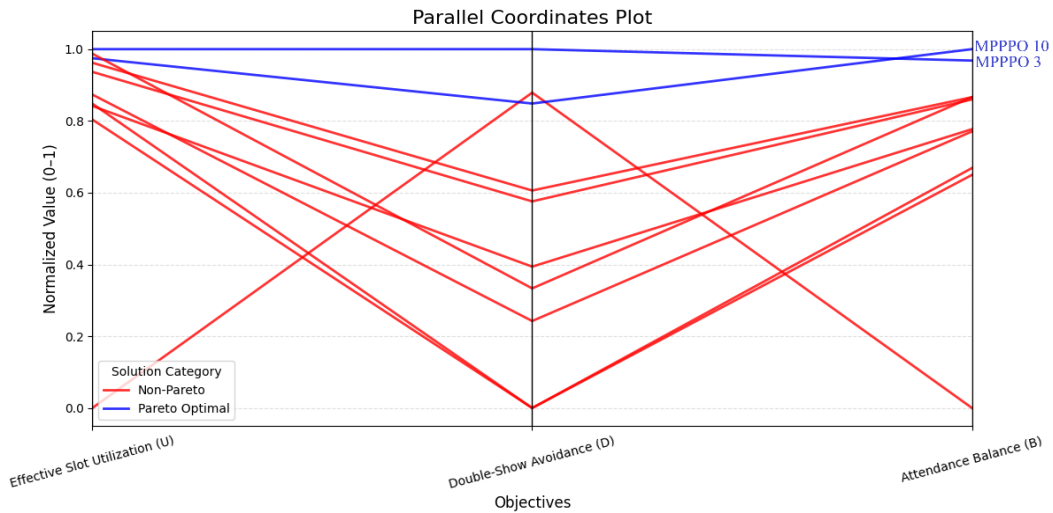

**Figure 8. Parallel Plot of 3 Objectives.** The figure visualizes trade-offs among objective slot utilization, double-show avoidance, and attendance balance across the ten trained MPPPO policies, highlighting MPPPO 3, and MPPPO 10 as approximate Pareto-optimal solutions.

Figure 9 further illustrates this effect by comparing MPPPO 1, MPPPO2, and MPPPO 3, each of which assigns full weight to a single objective. Despite prioritizing attendance balance, MPPPO 3 achieves higher utilization than MPPPO 1 and attains double-show avoidance comparable to MPPPO 2 ($D = 0.937$). These results indicate that emphasizing attendance balance can indirectly enhance performance in both utilization and double-show avoidance.

From an operational perspective, MPPPO 3 and MPPPO 10 offer balanced strategies for clinics seeking to improve effective slot utilization while controlling double-show risk. For settings that prioritize utilization, MPPPO 1, MPPPO 4, and MPPPO 8 are appropriate choices. For more conservative scheduling focused on minimizing double-shows, MPPPO 2 and MPPPO 10 provide the strongest avoidance performance.

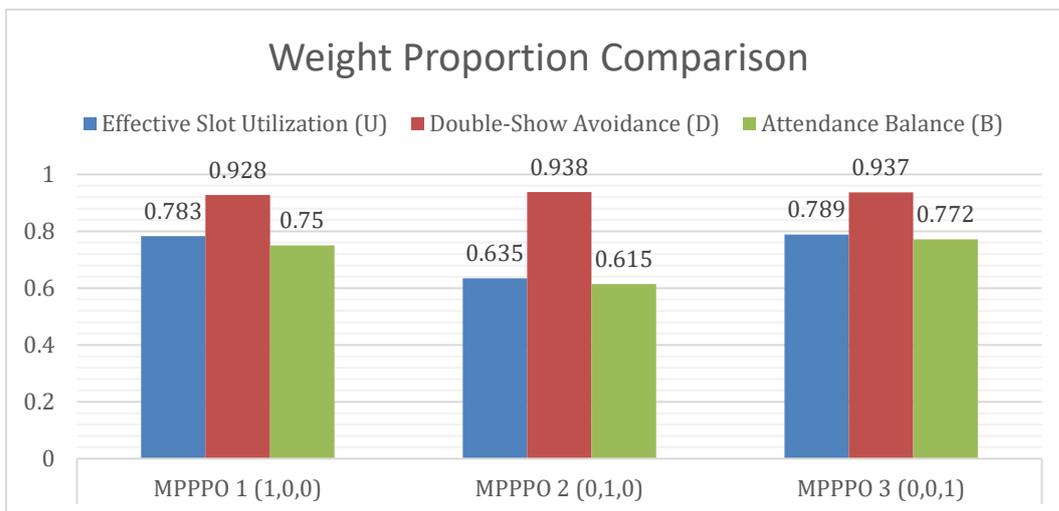

**Figure 9. Weight Proportion Comparison Result.** MPPPO 1, MPPPO 2, and MPPPO 3 represent policies optimized for utilization, double-show avoidance, and attendance balance, respectively, illustrating cross-objective performance effects.

#### 4.3.5 Sensitivity to No-Show Probability Perturbations

Robustness to prediction uncertainty is evaluated by perturbing predicted no-show probabilities by ±3% and ±5% during evaluation. Sensitivity is quantified as the relative change in average weighted reward. Figure 10 shows that performance remains stable under ±3% perturbations, with reward changes generally below 1%, indicating that the learned policies are robust to moderate prediction errors. Some policies exhibit slight improvements under small perturbations. For example, MPPPO 2 achieves a higher reward when probabilities are reduced by 3% suggesting that mild underestimation can lead to more conservative booking choices that better match realized attendance. Similarly, MPPPO 2, 3, 4, and 9 show modest gains under a +3% increase, which may occur when a small upward shift encourages additional double-booking in high-risk slots, improving utilization without substantially increasing double-show events.

In contrast, larger perturbations of ±5% lead to more pronounced performance degradation. Overestimation of no-show probability causes greater performance loss than underestimation, as it induces overly aggressive double-booking and increases double-show occurrences. Overall, these results demonstrate that the learned policies are robust to moderate prediction errors but sensitive to substantial overestimation of no-show risk.

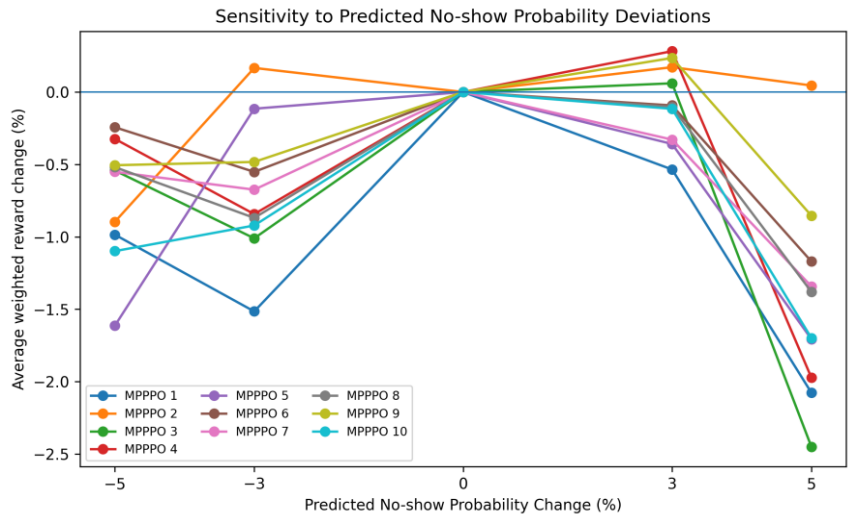

**Figure 10** Sensitivity to Predicted No-Show Probability Deviations. Reward sensitivity measured as relative change in average weighted reward under ±3% and ±5% perturbations of predicted no-show probabilities.

## 5. Conclusion

This study proposes an adaptive double-booking framework that integrates individualized no-show prediction with multi-objective RL for outpatient appointment scheduling. By combining MHASRF-based no-show probabilities estimates with an MPPPO algorithm, the framework learns data-driven booking decisions that balance effective slot utilization, overbooking risk, and attendance balance within a unified decision-making process. The multi-policy design enables schedulers to select or switch policies according to operational conditions, such as changing no-show patterns or demand fluctuations, thereby enhancing adaptability and robustness in practice.

Embedding patient-specific no-show probabilities into the RL state representation enables context-aware scheduling decisions that move beyond static or population-level assumptions. As a result, the agent applies double-booking selectively when the predicted absence risk is high, rather than relying on fixed rules. Simulation results demonstrate that the proposed approach outperforms both single-booking and fixed-threshold double-booking strategies, achieving higher effective slot utilization while maintaining low double-show risk.

The results also highlight that the attendance balance mediates the trade-off between utilization and congestion by encouraging expected attendance to remain close to one patient per slot. Policies that emphasize this objective achieve strong performance across all metrics, suggesting that attendance balance serves as an effective surrogate for overall scheduling quality. This finding provides practical guidance for designing reward structures in learning-based scheduling systems.

Interpretability analysis using SHAP confirms that the learned policies follow intuitive and operationally sound patterns, favoring single-booking for lower-risk and lower-load situations, reflecting a preference for stability and conservative scheduling. In contrast, double-booking is primarily driven by elevated no-show risk and is applied selectively when workload is manageable and scheduling flexibility remains, consistent with its intended role and support trust in the model's decisions.

By learning multiple co-evolving policies through MPPPO with the MPCEM procedure, the framework identifies approximate Pareto solutions corresponding to different operational priorities. This flexibility enables adaptive and resilient outpatient scheduling. Overall, the proposed approach offers an effective, interpretable, and data-driven solution for managing appointment scheduling under uncertainty. Future work may extend the framework to larger patient volumes and more complex end-to-end scheduling environments, and incorporate downstream operational outcomes such as patient waiting time, staff utilization, and clinic flow. These extensions would further strengthen the applicability of adaptive learning approaches in real-world healthcare operations.

**Funding**

This research was supported by Basic Science Research Program through the National Research Foundation of Korea(NRF) funded by the Ministry of Education (No. RS-2023-00248913).